\documentclass[11pt]{article}

\usepackage[preprint]{acl}

\usepackage{times}
\usepackage{latexsym}

\usepackage[T1]{fontenc}

\usepackage[utf8]{inputenc}

\usepackage{microtype}
\usepackage[most]{tcolorbox}
\usepackage{inconsolata}

\usepackage{graphicx}
\usepackage{xspace}
\usepackage{amssymb}
\usepackage{booktabs}
\usepackage{amsmath}
\usepackage{subcaption}
\usepackage{multirow}
\usepackage{amsthm}
\usepackage{algorithm}
\usepackage{algorithmic}

\newcommand{\llmname}[1]{{\texttt{#1}}}

\def\method{GoAgent\xspace}
\title{GoAgent: Group-of-Agents Communication Topology Generation for LLM-based Multi-Agent Systems}

\author{
\textbf{Hongjiang Chen\textsuperscript{1}},
\textbf{Xin Zheng\textsuperscript{2}},
 \textbf{Yixin Liu\textsuperscript{3}},
 \textbf{Pengfei Jiao\textsuperscript{1}},
 \textbf{Shiyuan Li\textsuperscript{3}},
 \textbf{Huan Liu\textsuperscript{1}},
 \\
 \textbf{Zhidong Zhao\textsuperscript{1}},
 \textbf{Ziqi Xu\textsuperscript{2}},
 \textbf{Ibrahim Khalil\textsuperscript{2}},
 \textbf{Shirui Pan\textsuperscript{3}},
 \\
 \textsuperscript{1}School of Cyberspace, Hangzhou Dianzi University, Hangzhou, China,
\\
 \textsuperscript{2}School of Computing Technologies, RMIT University, Melbourne, Australia,
 \\
 \textsuperscript{3}School of Information and Communication Technology, Griffith University, Goldcoast, Australia
 \\
 \{\texttt{hchen}, \texttt{pjiao}, \texttt{huanliu}, \texttt{zhaozd}\}@hdu.edu.cn, 
 \{\texttt{xin.zheng2}, \texttt{ziqi.xu}, \texttt{ibrahim.khalil}\}@rmit.edu.au, \\
 \{\texttt{yixin.liu}, \texttt{shiyuan.li}, \texttt{s.pan}\}@griffith.edu.au
}

\begin{document}
\maketitle
\begin{abstract}
Large language model (LLM)-based multi-agent systems (MAS) have demonstrated exceptional capabilities in solving complex tasks, yet their effectiveness depends heavily on the underlying communication topology that coordinates agent interactions. Within these systems, successful problem-solving often necessitates task-specific group structures to divide and conquer subtasks.
However, most existing approaches generate communication topologies in a node-centric manner, leaving group structures to emerge implicitly from local connectivity decisions rather than modeling them explicitly, often leading to suboptimal coordination and unnecessary communication overhead.
To address this limitation, we propose \textbf{\method} (Group-of-Agents), a communication topology generation method that explicitly treats \emph{collaborative groups} as the atomic units of MAS construction. 
Specifically, \method first enumerates task-relevant candidate groups through an LLM and then autoregressively selects and connects these groups as atomic units to construct the final communication graph, jointly capturing intra-group cohesion and inter-group coordination. 
To mitigate communication redundancy and noise propagation inherent in expanding topologies, we further introduce a conditional information bottleneck (CIB) objective that compresses inter-group communication, preserving task-relevant signals while filtering out redundant historical noise. 
Extensive experiments on six benchmarks demonstrate the state-of-the-art performance of \method with 93.84\% average accuracy while reducing token consumption by about 17\%.
\end{abstract}

\section{Introduction}

Large language model (LLM)-based multi-agent systems (MAS) have become a promising approach for solving complex tasks~\cite{tran2025multi}, ranging from code generation~\cite{zhang2024codeagent} to mathematical reasoning~\cite{yao2023react} and multi-step decision-making~\cite{guo2026embodied}. The core of any MAS is its communication topology, a directed graph that governs how agents interact, share information, and coordinate their efforts~\cite{wu2024autogen,liang2024encouraging,wang2025beyond}. Recent studies show that the underlying organizational structure impacts the overall performance of MAS more than the capabilities of individual agents~\cite{qian2025scaling,rizvi2026benefits,zhou2026multiagent}. Consequently, designing communication topologies that ensure both effective problem-solving and efficient token usage has become a central challenge.

Early approaches to topology design relied on static, manually crafted structures, such as chains for sequential reasoning~\cite{zhang2024chain}, trees for hierarchical deliberation~\cite{zhou2024language}, and fully connected graphs for exhaustive debate~\cite{hu2024learning}. While effective in narrow domains, these fixed topologies lack the flexibility to adapt to diverse task requirements~\cite{liu2025graph}. To overcome this rigidity, subsequent research introduced deep learning techniques to automatically learn topologies based on predefined templates. Methods like AgentPrune~\cite{zhang2025cut}, AgentDropout~\cite{wang2025agentdropout}, G-Designer~\cite{zhang2025g}, and GTD~\cite{jiang2025dynamic} take a dense template graph and optimize it by pruning or reweighting edges for specific tasks. However, these template-based methods remain bounded by the initial agent sets and connectivity patterns. Most recently, ARG-Designer~\cite{li2025assemble} moved beyond these template constraints by proposing an autoregressive approach that dynamically generates the collaboration graph from scratch, sequentially adding agents, assigning roles, and establishing connections.

\begin{figure}[!t]
  \centering
  \includegraphics[width=0.95\linewidth]{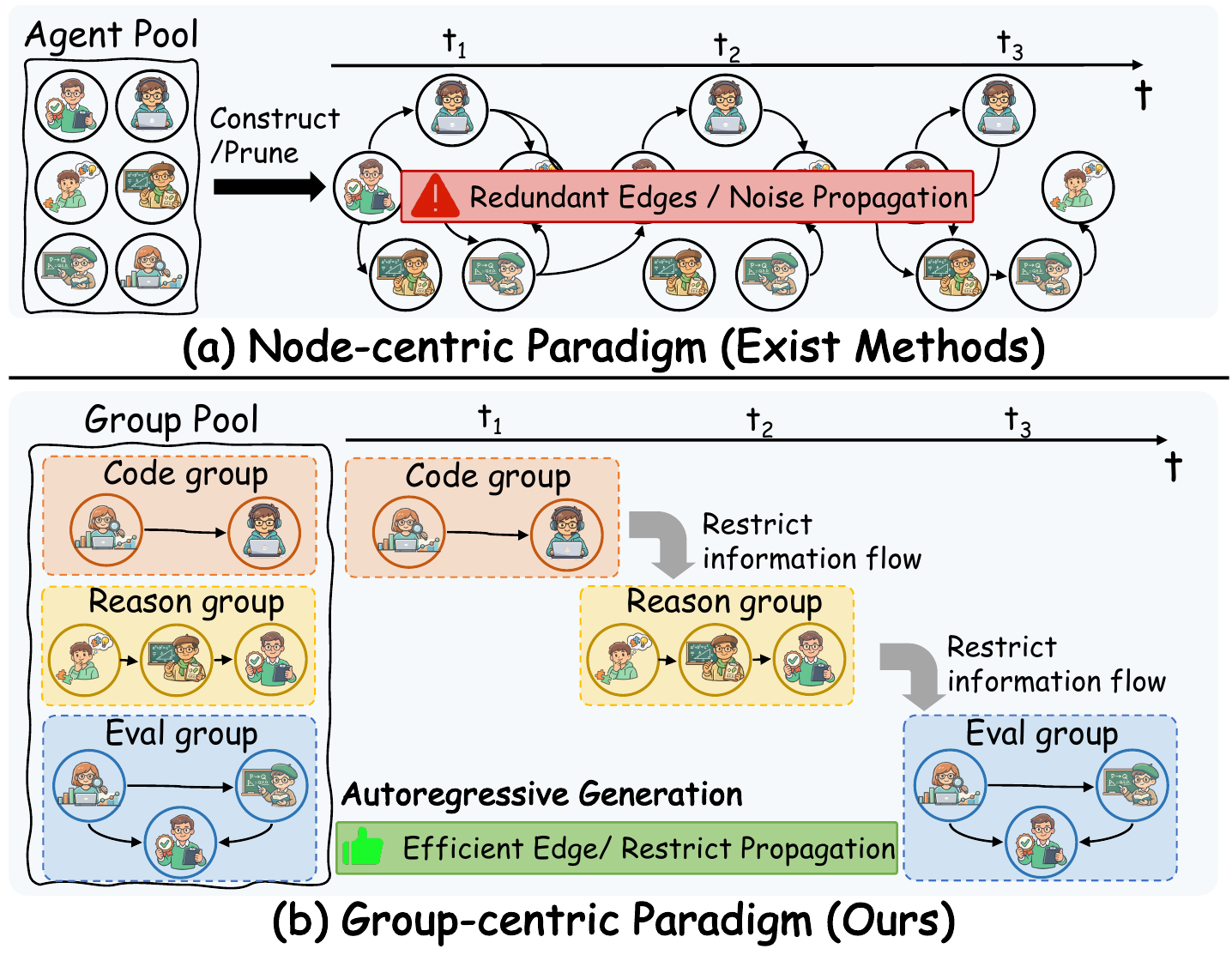}
  \caption{The comparison of two paradigms: (a) node-centric and (b) group-centric graph generation.}
  \label{fig:paradigm}
  \vspace{-2mm}
\end{figure}

Despite these advancements, existing topology generation methods share a fundamental flaw: they operate under a node-centric construction paradigm. As illustrated in Figure~\ref{fig:paradigm}(a), each agent is treated as an individual node and the topology is built by predicting agents and their connections step-by-step, resulting in isolated local decisions without a macroscopic view of the overall collaborative workflow.
This localized generation process introduces two critical limitations~\cite{barkoczi2016social}. \textit{First}, it hinders effective inter-agent coordination. Complex reasoning tasks naturally require a divide and conquer strategy executed by tightly coupled groups~\cite{betti2025dynamics}, such as a cohesive sub-team for problem solving consisting of a decomposer, a solver, and a verifier. Under the node-centric paradigm, these essential higher-order structures are not explicitly modeled; instead, they are expected to emerge implicitly from ad-hoc edge predictions, frequently resulting in disjointed workflows and suboptimal problem-solving~\cite{pescetelli2021modularity}. \textit{Second}, it exacerbates communication inefficiency and noise propagation~\cite{shen2025understanding,zhou2025guardian}. Without explicit group boundaries to structure information exchange, the generated graphs tend to form dense, unconstrained connections. This uncontrolled connectivity not only incurs substantial token overhead due to redundant message passing, but also allows task-irrelevant historical noise to accumulate and distract agents from critical signals. These limitations raise a critical question: 
\vspace{-2mm}
\begin{tcolorbox}[
colback=gray!5,
colframe=black,
boxrule=0.4pt,
left=2pt,
right=2pt,
top=2pt,
bottom=2pt
]
\textit{Can we move beyond the node-centric paradigm and design communication topologies using higher-level abstractions that explicitly capture and utilize collaborative group structures?}
\end{tcolorbox}

We answer this question by proposing \textbf{\method} (Group-of-Agents), a communication topology generation method that fundamentally shifts the paradigm by explicitly treating collaborative groups as the atomic units for construction (Figure~\ref{fig:paradigm}(b)). Given a specific task query, \method first prompts an LLM to enumerate a set of task-relevant candidate groups, where each group encapsulates a coherent cluster of expert roles tailored for a specific subtask. A learned autoregressive graph generation model then systematically selects and connects these predefined groups to construct the final communication topology. This group-level generation inherently preserves strong intra-group cohesion while explicitly directing inter-group coordination. To address the issue of communication inefficiency and noise propagation inherent in expanding topologies, \method incorporates a Conditional Information Bottleneck (CIB) mechanism into the graph generation process. 
Taking the specific task as the condition, the CIB layer actively compresses inter-group communication features, thereby retaining only the strictly necessary task-relevant signals and filtering out redundant historical noise.
Our main contributions are summarized as follows:
\begin{itemize}
  \item \textbf{New Paradigm.} We identify the inability of node-centric methods to explicitly model collaborative structures, proposing a group-centric paradigm that treats agent groups as atomic units of construction.
  \item \textbf{Novel Method.} We develop \method, an autoregressive model bridging LLM-derived agent groups with graph generation, and introduce CIB to adaptively compress redundant communications.
  \item \textbf{Experimental Validation.} Experiments on six benchmarks show that \method achieves state-of-the-art accuracy while reducing token consumption, establishing a new standard for efficient multi-agent system design.
\end{itemize}
\section{Preliminaries}
\subsection{Problem Formulation}
We study the problem of automatically constructing communication topologies for LLM-based MAS. 
Given a task query $\mathcal{Q}$, the objective is to generate a communication structure that coordinates a set of agents to collaboratively solve the task.

\paragraph{Agents.}
Let $\mathcal{V}=\{v_{1},\dots,v_{N}\}$ denote a set of $N$ candidate agents.
We instantiate each agent $v_i \in \mathcal{V}$ by assigning a role-specific system prompt (e.g., ``Math Solver'' or ``Code Reviewer'') to a shared large language model to define its expertise.
Given an input prompt $\mathcal{P}_i$, agent $v_i$ generates a natural language response $\mathcal{R}_i = v_i(\mathcal{P}_i)$.
Agents communicate via prompt composition: if the topology contains a directed edge from $v_i$ to $v_j$, the response $\mathcal{R}_i$ is appended to the input prompt of $v_j$ as context, facilitating information flow across the system.

\paragraph{Collaborative Groups.}
To capture structured cooperation among agents, we introduce collaborative groups as higher-order coordination units. 
Each group represents a structured collaboration pattern among a subset of agents and is defined as
$M_i = (\mathcal{S}_i, \mathcal{E}(\mathcal{S}_i))$, where $\mathcal{S}_i\subseteq\mathcal{V}$ denotes the subset of agents participating in group $M_i$, and $\mathcal{E}(\mathcal{S}_i)$ denotes the predefined communication relations (e.g., fully connected or chain) among agents within the group.
Let $\mathcal{M}=\{M_1,\dots,M_K\}$ denote the set of collaborative groups.

\paragraph{Group-Level Communication Graph.}
Based on these collaborative groups, we construct a group-level communication graph
\begin{equation}
\label{eq:graph}
\mathcal{G}=(\mathcal{M},\mathcal{E}),
\end{equation}
where $\mathcal{E}$ represents communication dependencies between groups. 
Each edge $e_{ij}\in\mathcal{E}$ indicates that information produced by group $M_i$ can be transmitted to group $M_j$. 
The final communication topology among agents is induced jointly by the intra-group relations $\mathcal{E}(\mathcal{S}_i)$ and the inter-group dependencies $\mathcal{E}$.

\paragraph{Topology Generation.}
Given a task query $\mathcal{Q}$, the goal is to generate the group-level communication graph $\mathcal{G}$. 
Groups are generated sequentially, and each newly generated group may establish connections with previously constructed groups. 
Formally, the generation distribution is defined as
\begin{equation}
\label{eq:generation}
\resizebox{.85\hsize}{!}{$
P(\mathcal{G}|\mathcal{Q})
=
\prod_{i=1}^{|\mathcal{M}|}
P(M_i| \mathcal{G}_{<i},\mathcal{Q})
\prod_{j<i}
P(e_{ji}| M_i,\mathcal{G}_{<i},\mathcal{Q}),$}
\end{equation}
where $\mathcal{G}_{<i}$ denotes the partial graph consisting of the first $i-1$ generated groups. 
The first term models collaborative group selection, while the second term predicts communication edges between the newly generated group and previously constructed groups.

\subsection{Information Bottleneck}
The information bottleneck (IB) principle~\cite{tishby2000information} seeks a compressed representation $\tilde{X}$ of signals $X$ that preserves the maximum information relevant to a target variable $Y$. The standard IB objective is formulated as minimizing:
\begin{equation}
\mathcal{L}_{\text{IB}} = -I(\tilde{X};Y) + \beta I(\tilde{X};X),
\end{equation}
where $\beta$ serves as a Lagrange multiplier that balances the trade-off between accurately predicting $Y$ and effectively compressing $X$.

While standard IB compresses information unconditionally, it can be extended to the CIB~\cite{gondek2003conditional} by introducing a condition variable $Z$. CIB seeks to extract $\tilde{X}$ that is maximally informative about $Y$ while compressing $X$, all conditioned on $Z$:
\begin{equation}
\min \mathcal{L}_{\text{CIB}} =  -I(\tilde{X};Y|Z) + \beta I(\tilde{X};X|Z).
\label{eq:cib}
\end{equation}
By conditioning on $Z$, CIB ensures that the compression selectively filters out noise while preserving the specific structural or semantic patterns dictated by the condition $Z$.
\begin{figure*}[!t]
  \centering
  \includegraphics[width=0.95\textwidth]{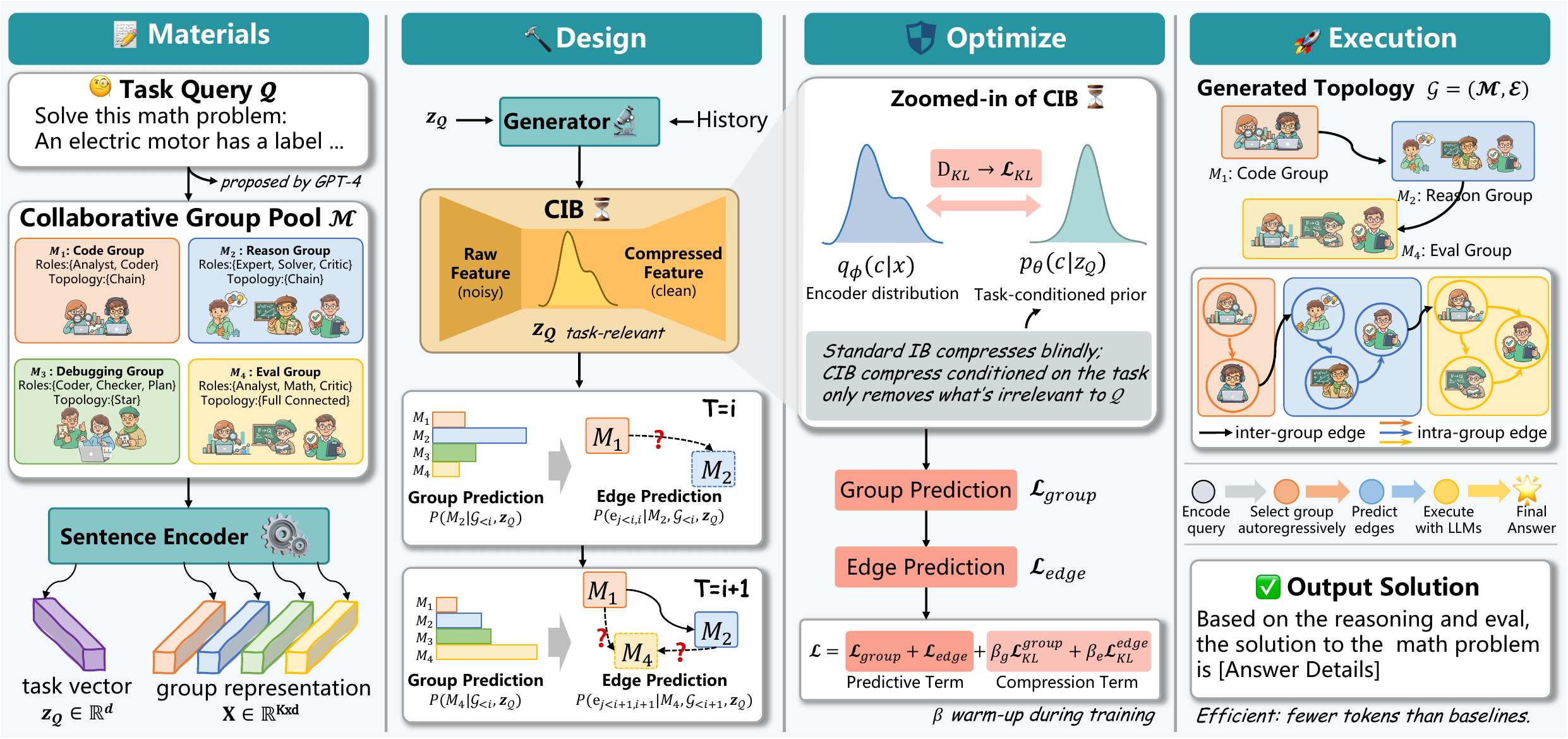}
  \caption{Overview of the proposed \method.}
  \label{fig:model}
\end{figure*}

\section{Methodology}
The primary goal of \method is to generate a task-specific, group-level communication topology for MAS. Unlike existing node-centric approaches that predict individual agents step-by-step, we introduce a group-centric paradigm where high-level collaborative groups serve as the generative atomic units. This macroscopic approach significantly reduces the search space and aligns better with human-like organizational structures.

As formulated in Eq.~\eqref{eq:generation}, \method factorizes the generation of topology $\mathcal{G}$ into an autoregressive sequence of group selections and edge predictions. The pipeline consists of four phases: (1) encoding the task query and discovering candidate groups; (2) autoregressively generating groups and their communication edges; (3) applying a CIB layer to compress historical communication signals; and (4) supervised training on curated ground-truth trajectories. The pseudo-code is provided in Appendix~\ref{sec:algorithm}.

\subsection{Task Encoding and Group Discovery}
\label{sec:task_enc}
We first map unstructured text, including both the task requirements and the agent descriptions, into a continuous vector space.

\paragraph{Task Encoding.} Given a task query $\mathcal{Q}$, we encode it into a global task representation $\mathbf{z}_{\mathcal{Q}} \in \mathbb{R}^d$ using a pre-trained sentence encoder followed by a feed-forward network (FFN):
\begin{equation}
    \mathbf{z}_{\mathcal{Q}} = \text{FFN}(\text{SentenceEncoder}(\mathcal{Q})).
\end{equation}
This representation $\mathbf{z}_{\mathcal{Q}}$ serves as the global condition guiding all subsequent generation steps.

\paragraph{Collaborative Group Discovery.} Instead of selecting individual agent instances, \method samples from a predefined pool of $K$ task-relevant collaborative groups. To construct this candidate space, we prompt a large language model (e.g., GPT-4) with domain-specific instructions to propose $K$ expert groups (e.g., $K=16$). Each group $M_i$ is defined by a structured schema: \texttt{(Name, Expertise, Roles, Intra-Topology)}. For example, a group might be defined as \texttt{Name}: ``Code Debugging Group'', \texttt{Expertise}: ``Identifies and fixes logical errors in code'', \texttt{Roles}: ``[Code Reviewer, Syntax Checker, Logic Validator]'', and \texttt{Intra-Topology}: ``Sequential Pipeline''. The intra-group edges $\mathcal{E}(\mathcal{S}_i)$ are LLM-inducted during this discovery phase and serve as fixed templates (e.g., fully connected for brainstorming, or sequential for refinement). These templates are rigidly enforced whenever the group is instantiated, allowing the autoregressive generator to focus entirely on the higher-level inter-group dependencies $\mathcal{E}$. We encode the textual descriptions of these $K$ groups using the same sentence encoder to form a candidate embedding matrix $\mathbf{X} = [\mathbf{x}_1, \dots, \mathbf{x}_K] \in \mathbb{R}^{K \times d}$. To facilitate autoregressive generation, we augment $\mathbf{X}$ with \texttt{END} tokens, resulting in a $(K+1) \times d$ embedding matrix.

\subsection{Autoregressive Group-Centric Generation}

At each time step $t$, the model determines the next group $M_t$ to add and establishes directed edges from the previously generated groups $\mathcal{G}_{<t}$ to $M_t$.

\paragraph{Historical Aggregation.} The model must aggregate the structural and semantic information of the current graph state. We use a gated recurrent unit (GRU) to compress the sequence of previously generated group embeddings into a historical state $\mathbf{h}_{\text{his}}^{(t)}$:
\begin{equation}
    \mathbf{h}_{\text{his}}^{(t)} = \text{GRU}([\mathbf{x}_{M_1}, \dots, \mathbf{x}_{M_{t-1}}]).
\end{equation}
To ensure the generation remains aligned with the task, we employ a dynamic gate $g^{(t)}$ to fuse the historical context with the task representation $\mathbf{z}_{\mathcal{Q}}$:
\begin{align}
    \label{eq:gate}
    g^{(t)} &= \sigma\left(\frac{\mathbf{h}_{\text{his}}^{(t)} \cdot \mathbf{z}_{\mathcal{Q}}}{\sqrt{d}}\right), \\
    \label{eq:fusion}
    \mathbf{h}_{\text{comb}}^{(t)} &= (1 - g^{(t)}) \mathbf{h}_{\text{his}}^{(t)} + g^{(t)} \mathbf{z}_{\mathcal{Q}} + \mathbf{e}_{\text{pos}}^{(t)},
\end{align}
where $\mathbf{e}_{\text{pos}}^{(t)}$ is a learnable step embedding providing positional information.

\paragraph{Group Prediction.} The fused state $\mathbf{h}_{\text{comb}}^{(t)}$ passes through a dedicated GRU to produce the raw group feature $\mathbf{x}_{\text{group}}^{(t)}$. Rather than using this raw feature directly, we pass it through the CIB layer (detailed in Section~\ref{sec:cib}) to obtain a compressed, noise-filtered representation $\mathbf{c}_{\text{group}}^{(t)}$. The probability distribution over the candidate groups is then computed via scaled dot-product attention with the candidate matrix $\mathbf{X}$:
\begin{equation}
    \label{eq:group_pred}
    P(M_t | \mathcal{G}_{<t}, \mathbf{z}_{\mathcal{Q}}) = \text{Softmax}(\mathbf{c}_{\text{group}}^{(t)} \mathbf{X}^\top).
\end{equation}

\paragraph{Edge Prediction.} Once $M_t$ is selected, the model predicts incoming edges from all existing groups $M_i \in \mathcal{G}_{<t}$. For each candidate edge $(M_i, M_t)$, we concatenate the historical state of the source group, the new group's embedding, and the task representation to form the raw edge feature $\mathbf{x}_{\text{edge}}^{(i,t)} = [\mathbf{h}_{\text{comb}}^{(i)} \| \mathbf{x}_{M_t} \| \mathbf{z}_{\mathcal{Q}}]$. This feature is similarly compressed by the CIB layer into $\mathbf{c}_{\text{edge}}^{(i,t)}$. A binary classifier then predicts the edge existence probability:
\begin{equation}
    \label{eq:edge_pred}
    P(e_{i,t} = 1 | M_i, \mathcal{G}_{<t}, \mathbf{z}_{\mathcal{Q}}) = \sigma(\text{MLP}(\mathbf{c}_{\text{edge}}^{(i,t)})).
\end{equation}

\subsection{Conditional Information Bottleneck}
\label{sec:cib}
As the communication graph grows, the raw historical features $\mathbf{x}$ (i.e., $\mathbf{x}_{\text{group}}^{(t)}$ or $\mathbf{x}_{\text{edge}}^{(i,t)}$) inevitably accumulate task-irrelevant signals, such as spurious group co-occurrences. To prevent this noise from propagating to the target prediction $y$ (the next group $M_t$ or edge $e_{i,t}$), we introduce a CIB layer. Following Eq.~\eqref{eq:cib}, we instantiate the condition variable as the global task representation $\mathbf{z}_{\mathcal{Q}}$. While primarily motivated by inter-group noise filtering, we apply CIB to both group and edge predictions for architectural uniformity. The CIB extracts a compressed latent representation $\mathbf{c}$ from $\mathbf{x}$ by optimizing the following objective:
\begin{equation}
    \label{eq:cib_overall}
    \min \mathcal{L}_{\text{CIB}} = \underbrace{- I(\mathbf{c};y|\mathbf{z}_{\mathcal{Q}})}_{\text{Predictive Term}} + \beta \underbrace{I(\mathbf{x};\mathbf{c}|\mathbf{z}_{\mathcal{Q}})}_{\text{Compression Term}}
\end{equation}
Because directly computing mutual information is intractable, we derive variational upper bounds to operationalize both terms.

\paragraph{Minimizing Predictive Term.} 
The first term ensures $\mathbf{c}$ retains sufficient information to predict $y$. We minimize its variational upper bound using a decoder $p_{\psi}$:
\begin{equation}
  \resizebox{.85\hsize}{!}{$- I(\mathbf{c};y|\mathbf{z}_{\mathcal{Q}}) \leq \mathbb{E}_{\mathbf{c} \sim q_{\phi}(\mathbf{c}|\mathbf{x})} [-\log p_{\psi}(y|\mathbf{c}, \mathbf{z}_{\mathcal{Q}})]
  \triangleq \mathcal{L}_{\text{task}}\,.$}
\end{equation}
In practice, because our architecture explicitly fuses the task query $\mathbf{z}_{\mathcal{Q}}$ into the historical state $\mathbf{x}$ prior to compression, we assume the variables form a Markov chain $\mathbf{z}_{\mathcal{Q}} \rightarrow \mathbf{x} \rightarrow \mathbf{c} \rightarrow y$. Under this assumption, $\mathbf{c}$ acts as a sufficient statistic for predicting $y$, rendering $y$ conditionally independent of $\mathbf{z}_{\mathcal{Q}}$ given $\mathbf{c}$. Consequently, we efficiently parameterize the decoder as $p_{\psi}(y|\mathbf{c})$, which serves as a valid approximation of $p_{\psi}(y|\mathbf{c}, \mathbf{z}_{\mathcal{Q}})$ provided that the predictive term successfully forces $\mathbf{c}$ to retain the task-relevant signals.

\paragraph{Minimizing Compression Term.} 
The second term restricts the flow of noisy historical information. Unlike standard Variational IB~\cite{alemi2017deep} that uses a fixed standard normal prior, we introduce a task-conditioned prior $p_{\theta}(\mathbf{c}|\mathbf{z}_{\mathcal{Q}})$. This conditional prior holds under the premise that different task types (e.g., math reasoning vs. code generation) inherently require different baseline topological structures, meaning the latent space of valid communication graphs is fundamentally task-dependent. We parameterize both the encoder $q_{\phi}(\mathbf{c}|\mathbf{x})$ and the conditional prior $p_{\theta}(\mathbf{c}|\mathbf{z}_{\mathcal{Q}})$ as multivariate Gaussians with diagonal covariances via MLPs. The mutual information is upper-bounded by their Kullback-Leibler (KL) divergence:
\begin{equation}
  \resizebox{.85\hsize}{!}{$I(\mathbf{x};\mathbf{c}|\mathbf{z}_{\mathcal{Q}}) \leq \mathbb{E}_{\mathbf{x},\mathbf{z}_{\mathcal{Q}}} \left[ D_{\text{KL}}(q_{\phi}(\mathbf{c} | \mathbf{x}) \parallel p_{\theta}(\mathbf{c} | \mathbf{z}_{\mathcal{Q}})) \right] \triangleq \mathcal{L}_{\text{KL}}\,.$}
\end{equation}
During training, we sample $\mathbf{c}$ using the reparameterization trick: 
\begin{equation}
    \label{eq:reparam}
    \mathbf{c} = \mu_{\phi}(\mathbf{x}) + \sigma_{\phi}(\mathbf{x}) \odot \epsilon, \quad \epsilon \sim \mathcal{N}(\mathbf{0}, \mathbf{I}).
\end{equation}

\paragraph{Tractable Objective.} 
By combining these bounds, we obtain an end-to-end optimizable loss function:
\begin{equation}
    \mathcal{L}_{\text{CIB}} \leq \mathcal{L}_{\text{task}} + \beta \mathcal{L}_{\text{KL}}.
\end{equation}
This formulation directly bridges to our final training objective (Section~\ref{sec:training}), where $\mathcal{L}_{\text{task}}$ corresponds to the supervised cross-entropy losses, and $\mathcal{L}_{\text{KL}}$ acts as a task-guided regularizer to filter structural noise.

\subsection{Training and Inference Strategy}
\label{sec:training}
\paragraph{Data Construction.} The autoregressive generator requires ground-truth sequences of groups and edges for supervised training. Since manually authoring optimal task-graph pairs is infeasible, we curate a training dataset $\mathcal{D} = \{(\mathcal{Q}, \mathcal{G}^*)\}$ via an automated heuristic exploration process. For a given set of training queries, we sample diverse candidate topologies $\mathcal{G}$ by varying group combinations and edge densities. We then execute these graphs on the target tasks using the underlying LLMs. Graphs that successfully produce the correct final answer are collected. To encourage efficiency, we filter these successful graphs to retain the minimal viable topologies (i.e., those with fewer groups and edges), which serve as our positive ground-truth trajectories $\mathcal{G}^*$.

\paragraph{Training Objective.} We train the model end-to-end using Teacher Forcing on the curated trajectories in $\mathcal{D}$, avoiding the high variance typical of online reinforcement learning. The total loss $\mathcal{L}$ is the empirical realization of the CIB objective across all generation steps. Specifically, the task loss $\mathcal{L}_{\text{task}}$ is instantiated as the negative log-likelihood for group prediction ($\mathcal{L}_{\text{group}}$) and binary cross-entropy for edge prediction ($\mathcal{L}_{\text{edge}}$):
\begin{align}
    \label{eq:loss_group}
    \mathcal{L}_{\text{group}} &= - \mathbb{E}_{\mathcal{D}} \left[ \sum_{t} \log P(M_t^* | \mathcal{G}_{<t}^*, \mathbf{z}_{\mathcal{Q}}) \right], \\
    \label{eq:loss_edge}
    \mathcal{L}_{\text{edge}} &= - \mathbb{E}_{\mathcal{D}} \left[ \sum_{t} \sum_{i<t} \log P(e_{i,t}^* | M_i^*, \mathcal{G}_{<t}^*, \mathbf{z}_{\mathcal{Q}}) \right].
\end{align}
The KL divergence regularizer $\mathcal{L}_{\text{KL}}$ is applied at both the group and edge levels. The final joint objective is:
\begin{equation}
    \label{eq:loss_total}
    \mathcal{L} = \underbrace{\mathcal{L}_{\text{group}} + \mathcal{L}_{\text{edge}}}_{\text{Predictive Task}} + \underbrace{\beta_g \mathcal{L}_{\text{KL}}^{\text{group}} + \beta_e \mathcal{L}_{\text{KL}}^{\text{edge}}}_{\text{Information Compression}},
\end{equation}
where $\beta_g$ and $\beta_e$ control the bottleneck strength. To prevent overly strong compression before the generator learns meaningful structural patterns, we employ a KL warm-up strategy, gradually increasing the $\beta$ weights during early training epochs.

\paragraph{Inference Procedure.} During inference, given a new task query $\mathcal{Q}$, the trained \method model generates a group-level graph autoregressively. The process, detailed in Algorithm~\ref{alg:generation}, starts with the encoded task embedding and enters a generation loop. At each step, it selects a new group using the deterministic mean to bypass stochastic sampling. If the selected group is not the special \texttt{END} token, the model then samples the existence of incoming edges from all previously generated groups via a Bernoulli distribution. This loop continues until the \texttt{END} token is produced.

\section{Experiments}
\paragraph{Benchmarks.}
Following~\cite{zhang2025g}, we evaluate our approach on general reasoning using MMLU~\cite{hendrycks2021measuring}, mathematical reasoning using GSM8K \cite{cobbe2021training}, MultiArith~\cite{roy2015solving}, SVAMP~\cite{patel2021nlp}, and AQuA~\cite{ling2017program}, as well as code generation using HumanEval~\cite{chen2021evaluating}. Details are provided in Appendix~\ref{sec:dataset_statistic}.

\paragraph{Baselines.}
We compare \method against three types of methods. The single-agent baselines include Vanilla (direct reasoning), Chain-of-Thought (CoT)~\cite{wei2022chain} and Self-Consistency (SC)~\cite{wang2023self}. For MAS with fixed topologies, we evaluate Chain, Tree, Complete graph, Random graph~\cite{qian2025scaling}, and LLM-Debate~\cite{du2024improving}. For MAS with learnable topologies, we include AgentPrune \cite{zhang2025cut}, AgentDropout~\cite{wang2025agentdropout}, G-Designer \cite{zhang2025g}, EIB-LEARNER \cite{shen2025understanding}, and ARG-Designer \cite{li2025assemble}. Further baseline details are in Appendix~\ref{sec:baseline_methods}.
\paragraph{Implementation Details.}
We access GPT models via the OpenAI API, primarily using \llmname{gpt-4o-2024-08-06}. A summarizer agent aggregates the dialogue history to produce the final solution, with the maximum interaction round set to $T = 3$ across all baselines and experiments. The group encoder uses \llmname{all-MiniLM-L6-v2}~\cite{wang2020minilm} with an embedding dimension of $d = 384$. Following standard LLM-MAS configurations~\cite{zhang2025g,li2025assemble}, we provide explicit agent profiles for multi-agent methods and use GPT-4 to generate this profile pool. For the CIB hyperparameters $\beta_g$ and $\beta_e$ (analyzed in Appendix~\ref{sec:parameter_sensitivity_analysis}), we apply a linear warm-up strategy over the first $E_{\text{warm}} = 10$ training epochs. For all datasets, we train the model using only $B \in \{40, 60\}$ queries. Additional implementation details are in Appendix~\ref{sec:implementation_details}.

\begin{table*}[!ht]
\centering
\caption{Overall performance comparison on six benchmarks (\%). The best results are highlighted in \textbf{bold}.}
\label{tab:main_results}
\resizebox{\textwidth}{!}{
\begin{tabular}{l|cccccc|c}
\toprule
\textbf{Method} & \textbf{MMLU} & \textbf{GSM8K} & \textbf{AQuA} & \textbf{MultiArith} & \textbf{SVAMP} & \textbf{HumanEval} & \textbf{Avg.} \\
\midrule
Vanilla  & 80.39  & 82.30  & 71.06  & 93.09  & 86.55  & 71.39  & 80.80 \\
\midrule
CoT  & 81.69 {\scriptsize$\uparrow 1.30$} & 86.50 {\scriptsize$\uparrow 4.20$} & 73.58 {\scriptsize$\uparrow 2.52$} & 93.25 {\scriptsize$\uparrow 0.16$} & 87.36 {\scriptsize$\uparrow 0.81$} & 74.67 {\scriptsize$\uparrow 3.28$} & 82.84 {\scriptsize$\uparrow 2.05$} \\

SC & 83.66 {\scriptsize$\uparrow 3.27$} & 81.60 {\scriptsize$\downarrow 0.70$} & 75.63 {\scriptsize$\uparrow 4.57$} & 94.12 {\scriptsize$\uparrow 1.03$} & 88.59 {\scriptsize$\uparrow 2.04$} & 79.83 {\scriptsize$\uparrow 8.44$} & 83.91 {\scriptsize$\uparrow 3.11$} \\
\midrule
Chain  & 83.01 {\scriptsize$\uparrow 2.62$} & 88.30 {\scriptsize$\uparrow 6.00$} & 74.05 {\scriptsize$\uparrow 2.99$} & 93.27 {\scriptsize$\uparrow 0.18$} & 87.17 {\scriptsize$\uparrow 0.62$} & 81.37 {\scriptsize$\uparrow 9.98$} & 84.53 {\scriptsize$\uparrow 3.73$} \\

Tree  & 81.04 {\scriptsize$\uparrow 0.65$} & 85.20 {\scriptsize$\uparrow 2.90$} & 71.23 {\scriptsize$\uparrow 0.17$} & 93.68 {\scriptsize$\uparrow 0.59$} & 88.91 {\scriptsize$\uparrow 2.36$} & 80.53 {\scriptsize$\uparrow 9.14$} & 83.43 {\scriptsize$\uparrow 2.64$} \\

Complete  & 82.35 {\scriptsize$\uparrow 1.96$} & 80.10 {\scriptsize$\downarrow 2.20$} & 72.95 {\scriptsize$\uparrow 1.89$} & 94.53 {\scriptsize$\uparrow 1.44$} & 84.01 {\scriptsize$\downarrow 2.54$} & 79.03 {\scriptsize$\uparrow 7.64$} & 82.16 {\scriptsize$\uparrow 1.37$} \\

Random  & 84.31 {\scriptsize$\uparrow 3.92$} & 86.90 {\scriptsize$\uparrow 4.60$} & 76.48 {\scriptsize$\uparrow 5.42$} & 94.08 {\scriptsize$\uparrow 0.99$} & 87.54 {\scriptsize$\uparrow 0.99$} & 82.66 {\scriptsize$\uparrow 11.27$} & 85.33 {\scriptsize$\uparrow 4.53$} \\

LLM-Debate  & 84.96 {\scriptsize$\uparrow 4.57$} & 91.40 {\scriptsize$\uparrow 9.10$} & 77.65 {\scriptsize$\uparrow 6.59$} & 96.36 {\scriptsize$\uparrow 3.27$} & 90.11 {\scriptsize$\uparrow 3.56$} & 84.70 {\scriptsize$\uparrow 13.31$} & 87.53 {\scriptsize$\uparrow 6.73$} \\
\midrule
AgentPrune  & 85.07 {\scriptsize$\uparrow 4.68$} & 91.10 {\scriptsize$\uparrow 8.80$} & 80.51 {\scriptsize$\uparrow 9.45$} & 94.65 {\scriptsize$\uparrow 1.56$} & 90.58 {\scriptsize$\uparrow 4.03$} & 86.75 {\scriptsize$\uparrow 15.36$} & 88.11 {\scriptsize$\uparrow 7.31$} \\

AgentDropout  & 85.62 {\scriptsize$\uparrow 5.23$} & 91.70 {\scriptsize$\uparrow 9.40$} & 80.94 {\scriptsize$\uparrow 9.88$} & 95.60 {\scriptsize$\uparrow 2.51$} & 91.04 {\scriptsize$\uparrow 4.49$} & 85.98 {\scriptsize$\uparrow 14.59$} & 88.48 {\scriptsize$\uparrow 7.68$} \\

G-Designer  & 86.92 {\scriptsize$\uparrow 6.53$} & 93.80 {\scriptsize$\uparrow 11.50$} & 81.60 {\scriptsize$\uparrow 10.54$} & 96.50 {\scriptsize$\uparrow 3.41$} & 93.10 {\scriptsize$\uparrow 6.55$} & 88.33 {\scriptsize$\uparrow 16.94$} & 90.04 {\scriptsize$\uparrow 9.25$} \\

EIB-LEARNER  & 88.90 {\scriptsize$\uparrow 8.51$} & 95.20 {\scriptsize$\uparrow 12.90$} & 83.46 {\scriptsize$\uparrow 12.40$} & 96.83 {\scriptsize$\uparrow 3.74$} & 94.70 {\scriptsize$\uparrow 8.15$} & 89.15 {\scriptsize$\uparrow 17.76$} & 91.37 {\scriptsize$\uparrow 10.58$} \\

ARG-Designer  & 89.54 {\scriptsize$\uparrow 9.15$} & 94.37 {\scriptsize$\uparrow 12.07$} & 85.51 {\scriptsize$\uparrow 14.45$} & 98.93 {\scriptsize$\uparrow 5.84$} & 95.63 {\scriptsize$\uparrow 9.08$} & 91.74 {\scriptsize$\uparrow 20.35$} & 92.62 {\scriptsize$\uparrow 11.82$} \\
\midrule
\textbf{\method~(Ours)}  & \textbf{91.50} {\scriptsize$\uparrow 11.11$} & \textbf{95.30} {\scriptsize$\uparrow 13.00$} & \textbf{86.45} {\scriptsize$\uparrow 15.39$} & \textbf{99.11} {\scriptsize$\uparrow 6.02$} & \textbf{96.46} {\scriptsize$\uparrow 9.91$} & \textbf{94.21} {\scriptsize$\uparrow 22.82$} & \textbf{93.84} {\scriptsize$\uparrow 13.04$} \\
\bottomrule
\end{tabular}}
\end{table*}

\subsection{Performance Comparison}
Table~\ref{tab:main_results} shows that \method achieves consistent state-of-the-art accuracy across all six benchmarks. Compared to node-centric autoregressive methods like ARG-Designer and template-based pruning methods like EIB-LEARNER, \method shows larger gains on complex reasoning tasks, such as MMLU (\textbf{1.96}\% $\uparrow$) and HumanEval (\textbf{2.47}\% $\uparrow$). These improvements on harder tasks support our core motivation: constructing topologies at the group level leverages cohesive expert clusters, such as a solver paired with a verifier. This avoids the disjointed workflows and missing inter-role connections common in node-centric generation.

\subsection{Ablation Study}

\begin{table}[!t]
    \centering
    \caption{Ablation study of \method\ components.} 
    \label{tab:ablation}
    \resizebox{\columnwidth}{!}{
    \begin{tabular}{l|ccc|c}
    \toprule
    \textbf{Model} & \textbf{MMLU} & \textbf{GSM8K} & \textbf{HumanEval} & \textbf{Avg.} \\
    \midrule
    Vanilla & 80.39 & 82.30 & 71.39 & 78.03 \\
    \textbf{\method} & \textbf{91.50} & \textbf{95.30} & \textbf{94.21} & \textbf{93.67} \\
    \midrule
    w/o Group & 88.89 & 93.75 & 91.74 & 91.46 \\
    w/o CIB & 88.23 & 93.96 & 92.56 & 91.58 \\
    w/o ALL & 86.92 & 91.16 & 90.56 & 89.55 \\
    \bottomrule
    \end{tabular}}
\end{table}

We conduct ablation studies to isolate the contributions of key components in \method, with results detailed in Table~\ref{tab:ablation}. We evaluate three variants: (1) the w/o CIB variant removes the Conditional Information Bottleneck layer, forcing the model to rely on uncompressed historical features; (2) the w/o Group variant replaces group-level candidates with individual agent roles, reverting to a node-centric generation paradigm; and (3) the w/o ALL variant removes both mechanisms.
Removing either component leads to clear performance drops. For instance, the w/o Group variant suffers a \textbf{2.61\%} accuracy decrease on MMLU, confirming that implicit coordination through node-level generation struggles with complex tasks. Similarly, the w/o CIB variant experiences a \textbf{3.27\%} drop on MMLU. Without noise compression, the system becomes vulnerable to irrelevant historical signals, which misguide the edge prediction module into forming spurious connections. Together, these findings show that group-level generation and structural noise filtering are highly complementary.

\begin{figure}[!ht]
\centering
\begin{subfigure}[t]{0.48\columnwidth}
\centering
\includegraphics[width=\linewidth]{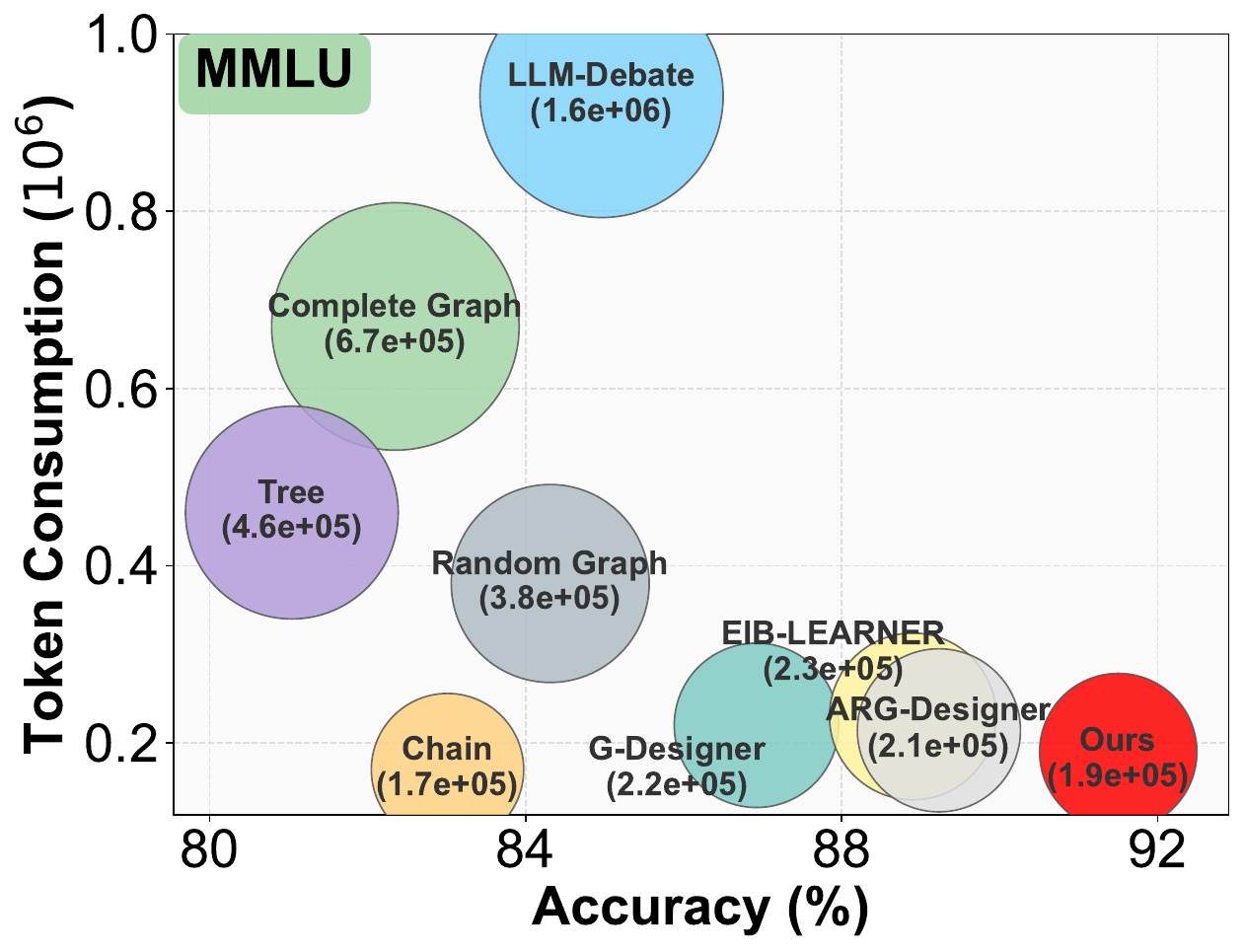}
\caption{Token cost on MMLU.}
\label{fig:token_mmlu}
\end{subfigure}
\hspace{-2mm}
\begin{subfigure}[t]{0.48\columnwidth}
\centering
\includegraphics[width=\linewidth]{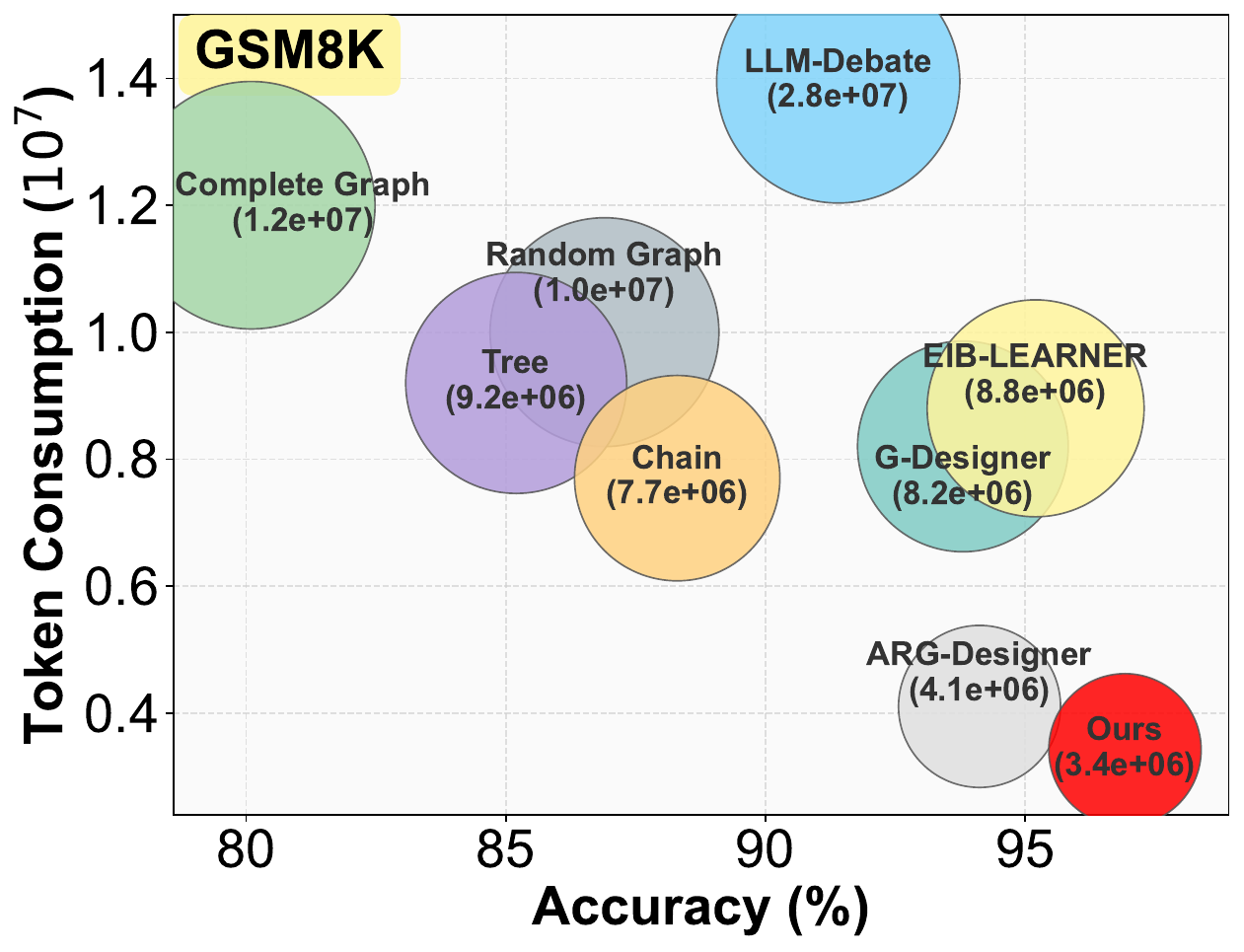}
\caption{Token cost on GSM8K.}
\label{fig:token_gsm8k}
\end{subfigure}
\begin{subfigure}[t]{\columnwidth}
  \centering
  \includegraphics[width=\linewidth]{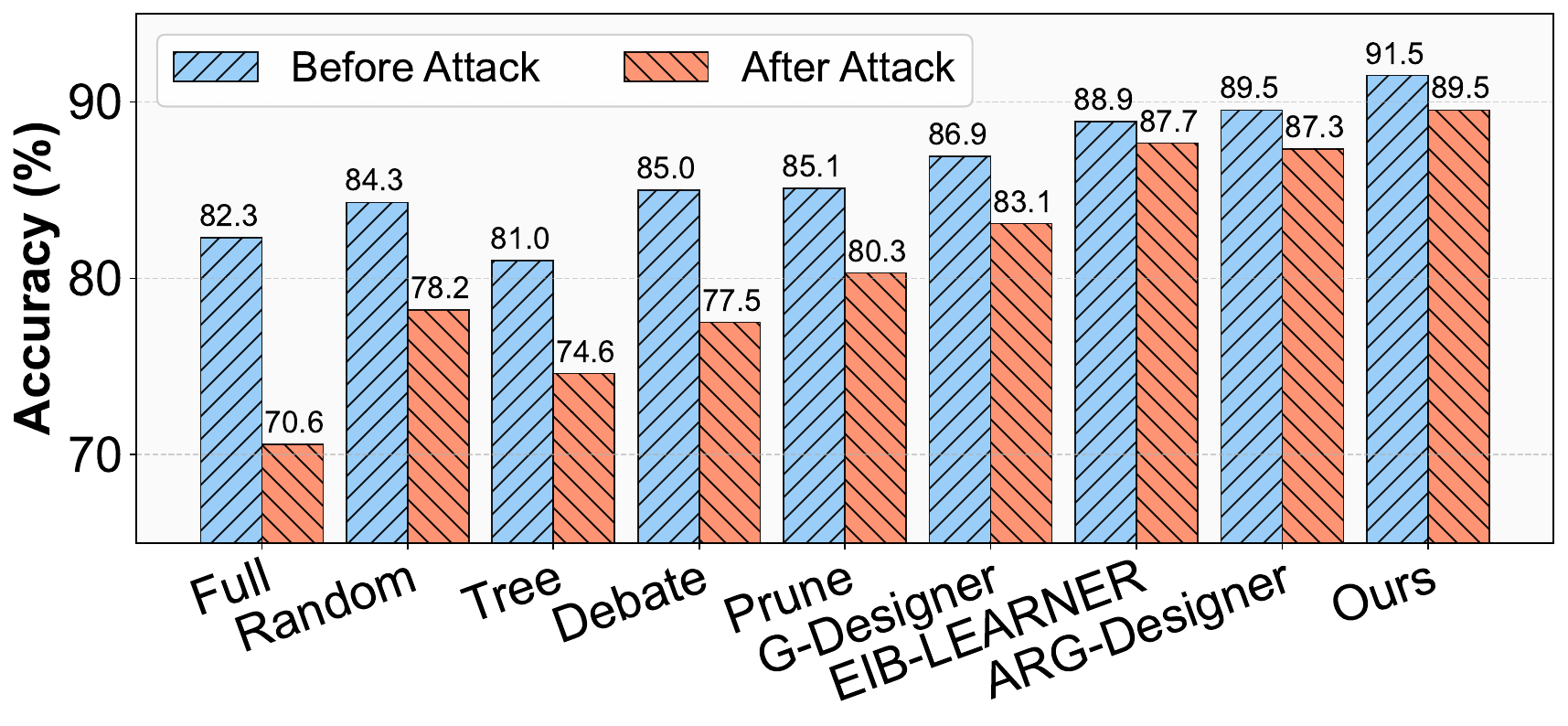}
  \caption{Robustness under attack.}
  \label{fig:robustness}
  \end{subfigure}
\caption{Experiments on token consumption and robustness against prompt injection attacks.}
\end{figure}

\subsection{Token Efficiency}
A key motivation for moving beyond node-centric generation is to reduce inference-time communication overhead. Figures~\ref{fig:token_mmlu} and \ref{fig:token_gsm8k} report the total LLM inference tokens on MMLU and GSM8K. Dense topologies like Complete graphs and LLM-Debate consume excessive tokens due to unrestricted information exchange. Although ARG-Designer learns sparse structures, its node-centric approach still yields unnecessary point-to-point connections. In contrast, \method reduces token consumption by \textbf{17\%} compared to the SOTA baseline, while maintaining high accuracy. By treating cohesive groups as atomic blocks and using the CIB layer to filter irrelevant inter-group edges, the model avoids redundant agent-level connections. This allows \method to maintain a token footprint comparable to simple trees while achieving the performance of highly expressive topologies.

\subsection{Robustness Analysis}
Following~\cite{zhuge2024gptswarm}, we simulate a system prompt attack by injecting adversarial prompts into a single agent. As illustrated in Figure~\ref{fig:robustness}, node-centric methods such as G-Designer and ARG-Designer suffer sharp performance drops under these attacks, because local errors rapidly propagate through their networks. In contrast, \method shows significantly better robustness, maintaining an accuracy of \textbf{89.54\%}. This resilience stems from the CIB layer, which acts as a filter by compressing incoming features conditioned on the task representation. Through this mechanism, the model learns to identify and discard signals that deviate from the task goal, thereby restricting the spread of localized hallucinations across group boundaries.

\begin{figure}[!ht]
  \centering
  \includegraphics[width=0.75\linewidth]{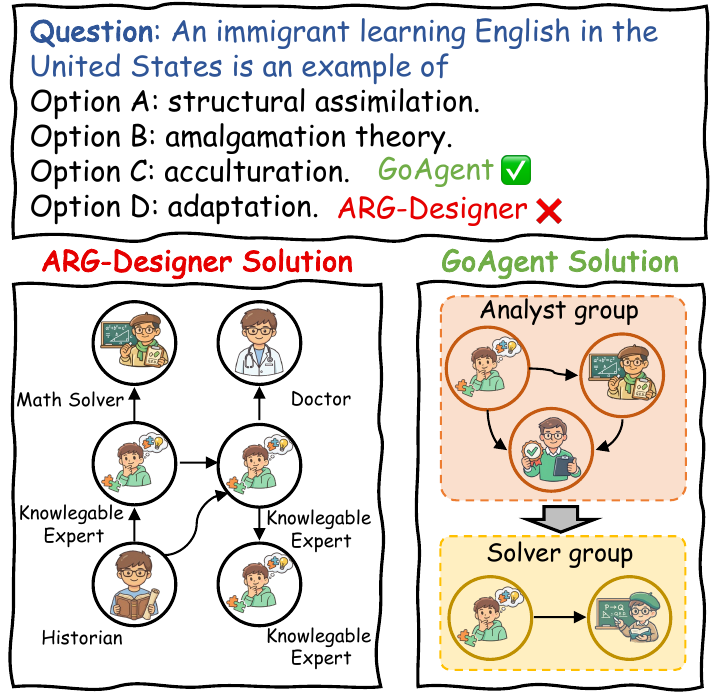}
  \caption{Cases of ARG-Designer and \method.}
  \label{fig:case}
  \vspace{-1mm}
\end{figure}

\subsection{Case Study}
To further illustrate the advantages of \method over learning-based baselines like ARG-Designer, we conduct a comparative case study on representative MMLU examples, shown in Figure~\ref{fig:case}. The main difference lies in the redundancy of composition and link generation. Without explicit higher-order constraints, ARG-Designer tends to connect redundant agents, and the density of its generated graphs increases sharply with task complexity. \method instead selects groups and inter-group links from an explicitly constructed pool, reducing unnecessary communication. Consequently, \method constructs more efficient collaboration graphs with fewer agents and messages, lowering token usage without sacrificing accuracy.

\section{Related Work}
\label{sec:related_work}
\paragraph{LLM-based Multi-Agent Systems.}
MAS coordinate diverse roles to solve complex tasks \cite{zhang2024codeagent,yao2023react,tran2025multi}, and their performance depends heavily on the underlying communication topology. Early architectures relied on static structures such as chains \cite{zhang2024chain,zhou2024language}. To enable adaptive communication, template-based methods like G-Designer, AgentPrune, and AgentDropout \cite{zhang2025g,zhang2025cut,wang2025agentdropout,shen2025understanding,jiang2025dynamic,wu2026st} optimize topologies by pruning predefined graphs. 
More recently, ARG-Designer \cite{li2025assemble} bypassed template constraints entirely by autoregressively constructing topologies from scratch. However, all these methods operate at the node level, treating individual agents as atomic units. This node-centric paradigm has difficulty capturing cohesive collaborative structures and frequently generates redundant edges.
\paragraph{Group-Aware Graph Generation.}
Node-centric graph generation often faces scalability and structural coherence challenges. To overcome these limitations, research in graph representation learning has explored higher-order paradigms. Hierarchical networks \cite{jin2020hierarchical,karami2023higen} explicitly model macroscopic subgraphs before establishing microscopic connections. Diffusion-based methods \cite{liu2024graph,xu2025modiff} incorporate global and local constraints to ensure structural validity, and scalable frameworks \cite{liu2023using,sarpe2024scalable} demonstrate that generating graphs through higher-order units improves both efficiency and fidelity. Drawing on these principles, \method introduces a group-centric paradigm to MAS. By using LLM-derived collaborative groups as atomic building blocks, it preserves intra-group cohesion and simplifies inter-group coordination.

\section{Conclusion}
In this paper, we present \method, a communication topology generation method for LLM-based multi-agent systems, driven by a group-centric design paradigm. By treating collaborative groups as atomic generative units and using a CIB to compress historical signals, \method adaptively forms task-specific topologies while suppressing task-irrelevant noise. Experiments across six benchmarks show that \method consistently achieves state-of-the-art accuracy, improves structural robustness, and reduces communication overhead.

\section*{Limitations}
Although our method demonstrates strong performance and efficiency across reasoning and coding benchmarks, several limitations remain. First, our approach relies on a predefined pool of collaborative groups generated offline by an LLM. While this significantly reduces the search space and ensures intra-group cohesion, it inherently constrains the model's flexibility; specifically, \method cannot dynamically synthesize entirely novel group structures or roles during inference if a task requires expertise outside the predefined pool. Second, our current evaluation primarily focuses on static reasoning tasks (e.g., MMLU, GSM8K). The effectiveness of group-centric topology generation in highly dynamic, interactive environments, including embodied AI or multi-agent reinforcement learning, remains to be thoroughly investigated in future research.

\section*{Ethics Statement}
Our research focuses on scientific questions and does not involve human subjects, animals, or environmentally sensitive materials. We foresee no ethical risks or conflicts of interest. We are committed to maintaining scientific integrity and ethics to ensure the validity and reliability of our findings.

\bibliography{ref}

\clearpage
\appendix

\section{More Related Work}
\label{sec:more_related_work}
\paragraph{Information Bottleneck in Graph Learning.}
The IB principle \cite{tishby2015deep,alemi2017deep} extracts minimal sufficient statistics for a task, balancing compression and prediction. In graph learning, IB optimizes communication efficiency by minimizing message entropy \cite{wang2020learning} or dynamically pruning redundant messages \cite{yuan2024dynamic}. Recently, IB has been applied to LLM-based MAS; for example, GUARDIAN \cite{zhou2025guardian} compresses temporal interaction graphs to mitigate hallucination and error propagation. Building on this, \method extends IB to the generative process of MAS topologies. We introduce a Conditional IB that dynamically conditions compression on the task representation, actively filtering out redundant historical communication noise as the topology expands.

\section{Algorithm and Complexity Analysis}
\label{sec:algorithm}
The complete training procedure of \method is summarized in Algorithm~\ref{alg:training}, and the inference procedure is detailed in Algorithm~\ref{alg:generation}. During inference, we bypass the stochastic sampling of the CIB layer and directly use the deterministic mean $\mathbf{c} = \mu_{\phi}(\mathbf{x})$ for both group and edge predictions.

\paragraph{Time Complexity.} 
Let $T$ be the total number of generated groups, $K$ be the candidate pool size, and $d$ be the hidden dimension. At each step $t$, the group prediction involves GRU state updates and attention over $K$ candidates, taking $\mathcal{O}(d^2 + K d)$ operations. The edge prediction evaluates connections from all $t-1$ previous groups, taking $\mathcal{O}(t \cdot d^2)$ operations. Summing over $T$ steps, the total time complexity for graph generation is $\mathcal{O}(T d^2 + T K d + T^2 d^2) = \mathcal{O}(T^2 d^2 + T K d)$. Because the generation is performed at the macroscopic group level rather than the microscopic agent level, both $T$ (typically $<3$) and $K$ (e.g., $16$) are extremely small. Consequently, the computational overhead of the generator is negligible compared to the inference cost of the underlying LLMs executing the task.

\paragraph{Space Complexity.}
The learnable parameters of the generator consist solely of lightweight GRUs, MLPs, and CIB projection layers, all bounded by $\mathcal{O}(d^2)$. The candidate group embeddings require $\mathcal{O}(K d)$ memory. Thus, the total space complexity is $\mathcal{O}(d^2 + K d)$, making the generator highly parameter-efficient and easy to integrate alongside large language models.

\begin{algorithm*}[t]
  \caption{Training Procedure of \method}
  \label{alg:training}
  \begin{algorithmic}[1]
  \REQUIRE Training dataset $\mathcal{D} = \{(\mathcal{Q}, \mathcal{G}^*)\}$ curated via heuristic exploration, Total epochs $E$, Warm-up epochs $E_{\text{warm}}$
  \ENSURE Optimized model parameters
  \FOR{epoch $e = 1$ \TO $E$}
      \STATE {\color{blue}\ttfamily{{/* Update CIB Bottleneck Strength */}}}
      \STATE Compute $\beta_g, \beta_e$ based on warm-up schedule (e.g., linear increase if $e < E_{\text{warm}}$)
      \FOR{each batch $(\mathcal{Q}, \mathcal{G}^*) \in \mathcal{D}$}
          \STATE Encode task query: $\mathbf{z}_{\mathcal{Q}} \leftarrow \text{TaskEncoder}(\mathcal{Q})$
          \STATE Initialize batch losses: $\mathcal{L}_{\text{group}} \leftarrow 0$, $\mathcal{L}_{\text{edge}} \leftarrow 0$, $\mathcal{L}_{\text{KL}}^{\text{group}} \leftarrow 0$, $\mathcal{L}_{\text{KL}}^{\text{edge}} \leftarrow 0$
          \FOR{step $t = 1$ \TO $|\mathcal{M}^*|$}
              \STATE {\color{blue}\ttfamily{/* Teacher Forcing with Ground-Truth History */}}
              \STATE Aggregate history $\mathbf{h}_{\text{his}}^{(t)}$ using ground-truth sequence $\mathcal{M}^*_{<t}$
              \STATE Fuse with task $\mathbf{z}_{\mathcal{Q}}$ to obtain $\mathbf{h}_{\text{comb}}^{(t)}$ via Eq.~\eqref{eq:gate} and Eq.~\eqref{eq:fusion}
              
              \STATE {\color{blue}\ttfamily{/* Group CIB \& Prediction Loss */}}
              \STATE Extract raw group feature $\mathbf{x}_{\text{group}}^{(t)}$
              \STATE Sample $\mathbf{c}_{\text{group}}^{(t)}$ using the reparameterization trick via Eq.~\eqref{eq:reparam}
              \STATE Accumulate $\mathcal{L}_{\text{group}}$ via Eq.~\eqref{eq:loss_group}
              \STATE Accumulate $\mathcal{L}_{\text{KL}}^{\text{group}}$ based on expected KL divergence
              
              \STATE {\color{blue}\ttfamily{/* Edge CIB \& Prediction Loss */}}
              \FOR{each existing group $M_i^* \in \mathcal{M}^*_{<t}$}
                  \STATE Construct raw edge feature $\mathbf{x}_{\text{edge}}^{(i,t)} = [\mathbf{h}_{\text{comb}}^{(i)} \| \mathbf{x}_{M_t^*} \| \mathbf{z}_{\mathcal{Q}}]$
                  \STATE Sample $\mathbf{c}_{\text{edge}}^{(i,t)}$ using the reparameterization trick via Eq.~\eqref{eq:reparam}
                  \STATE Compute $p_{i,t}$ via Eq.~\eqref{eq:edge_pred}
                  \STATE Accumulate $\mathcal{L}_{\text{edge}}$ via Eq.~\eqref{eq:loss_edge}
                  \STATE Accumulate $\mathcal{L}_{\text{KL}}^{\text{edge}}$ based on expected KL divergence
              \ENDFOR
          \ENDFOR
          \STATE {\color{blue}\ttfamily{/* Optimization */}}
          \STATE Compute total loss $\mathcal{L}$ via Eq.~\eqref{eq:loss_total}
          \STATE Update parameters using gradient descent on $\mathcal{L}$
      \ENDFOR
\ENDFOR
\RETURN Optimized parameters
\end{algorithmic}
\end{algorithm*}

\begin{algorithm*}[t]
\caption{Inference Procedure of \method}
\label{alg:generation}
\begin{algorithmic}[1]
\REQUIRE Task query $\mathcal{Q}$, Candidate group embeddings $\mathbf{X} \in \mathbb{R}^{(K+1) \times d}$
\ENSURE Group-level communication graph $\mathcal{G} = (\mathcal{M}, \mathcal{E})$
\STATE \textbf{Initialize:} $\mathcal{M} \leftarrow \emptyset$, $\mathcal{E} \leftarrow \emptyset$, $t \leftarrow 1$
\STATE Encode task query to global representation: $\mathbf{z}_{\mathcal{Q}} \leftarrow \text{TaskEncoder}(\mathcal{Q})$
\WHILE{True}
    \STATE {\color{blue}\ttfamily{/* 1. Historical Aggregation */}}
    \STATE Aggregate history $\mathbf{h}_{\text{his}}^{(t)} \leftarrow \text{GRU}([\mathbf{x}_{M_1}, \dots, \mathbf{x}_{M_{t-1}}])$
    \STATE Compute dynamic gate $g^{(t)}$ and fused state $\mathbf{h}_{\text{comb}}^{(t)}$ via Eq.~\eqref{eq:gate} and Eq.~\eqref{eq:fusion}
    
    \STATE {\color{blue}\ttfamily{/* 2. Group Prediction */}}
    \STATE Extract raw group feature $\mathbf{x}_{\text{group}}^{(t)}$ from $\mathbf{h}_{\text{comb}}^{(t)}$ via GRU
    \STATE Compress feature using deterministic CIB mean: $\mathbf{c}_{\text{group}}^{(t)} \leftarrow \mu_{\phi}(\mathbf{x}_{\text{group}}^{(t)})$
    \STATE Compute probability distribution over candidates $P(M_t)$ via Eq.~\eqref{eq:group_pred}
    \STATE Sample next group $M_t = \arg\max P(M_t)$  
    \IF{$M_t == \texttt{END}$}
        \STATE \textbf{break}
    \ENDIF
    \STATE Add group to graph: $\mathcal{M} \leftarrow \mathcal{M} \cup \{M_t\}$
    
    \STATE {\color{blue}\ttfamily{/* 3. Edge Prediction */}}
    \FOR{each existing group $M_i \in \mathcal{M}_{<t}$}
        \STATE Construct raw edge feature: $\mathbf{x}_{\text{edge}}^{(i,t)} = [\mathbf{h}_{\text{comb}}^{(i)} \| \mathbf{x}_{M_t} \| \mathbf{z}_{\mathcal{Q}}]$
        \STATE Compress feature using deterministic CIB mean: $\mathbf{c}_{\text{edge}}^{(i,t)} \leftarrow \mu_{\phi}(\mathbf{x}_{\text{edge}}^{(i,t)})$
        \STATE Predict edge probability $p_{i,t}$ via Eq.~\eqref{eq:edge_pred}
        \STATE Sample $a_{i,t} \sim \text{Bernoulli}(p_{i,t})$
        \IF{$a_{i,t} = 1$} 
            \STATE $\mathcal{E} \leftarrow \mathcal{E} \cup \{(M_i, M_t)\}$
        \ENDIF
    \ENDFOR
    \IF{no edge was added for $M_t$}
        \STATE $i^* \leftarrow \arg\max_{i} p_{i,t}$; \quad $\mathcal{E} \leftarrow \mathcal{E} \cup \{(M_{i^*}, M_t)\}$ 
    \ENDIF
    \STATE $t \leftarrow t + 1$
\ENDWHILE
\RETURN $\mathcal{G} = (\mathcal{M}, \mathcal{E})$
\end{algorithmic}
\end{algorithm*}

\section{Dataset Details}
\label{sec:dataset_statistic}
\begin{table*}[!ht]
  \centering
  \caption{Dataset descriptions and statistics.}
  \label{tab:dataset_stats}
  \begin{tabular}{l l l l c l}
  \toprule
  Category & Dataset & Answer Type & Metric & \#Test & License \\
  \midrule
  General reasoning & MMLU & Multi-choice & Acc. & 153 & MIT License \\
  \midrule
  \multirow{4}{*}{Math reasoning}
  & GSM8K      & Number       & Acc.   & 1,319 & MIT License \\
  & MultiArith & Number       & Acc.   & 600   & Unspecified \\
  & SVAMP      & Number       & Acc.   & 1,000 & MIT License \\
  & AQuA       & Multi-choice & Acc.   & 254   & Apache-2.0 \\
  \midrule
  Code generation & HumanEval & Code & Pass@1 & 164 & MIT License \\
  \bottomrule
  \end{tabular}
\end{table*}
We present the dataset statistics in Table~\ref{tab:dataset_stats}, following the same experimental setup as G-Designer~\cite{zhang2025g}.
We evaluate \method on six diverse benchmarks spanning multiple reasoning domains.

\section{Baseline Details}
\label{sec:baseline_methods}
\paragraph{Single-Agent Methods.}
We compare against three single-agent baselines: (1) \textbf{Vanilla}: Direct prompting without reasoning enhancement. (2) \textbf{Chain-of-Thought (CoT)}~\cite{wei2022chain}: Prompts the LLM to generate intermediate reasoning steps with ``Let's think step by step''. (3) \textbf{Self-Consistency (SC)}~\cite{wang2023self}: Samples five reasoning paths using CoT and selects the most consistent answer through majority voting.

\paragraph{Fixed Multi-Agent Topologies.}
We evaluate five hand-crafted topologies:
(1) \textbf{Chain}: Sequential linear chain $A_1 \rightarrow A_2 \rightarrow \cdots \rightarrow A_n$. (2) \textbf{Tree}: Hierarchical structure with multiple layers culminating in a root agent. (3) \textbf{Complete}: Fully connected graph enabling maximum information sharing. (4) \textbf{Random}: Randomly generated topology with random role assignments. (5) \textbf{LLM-Debate}: Debate-based topology where agents critique each other's reasoning over multiple rounds.

\paragraph{Learning-Based Topology Design.}
We compare against five state-of-the-art learning-based methods, each trained separately per dataset:
(1) \textbf{AgentPrune}~\cite{zhang2025cut}: Identifies and addresses communication redundancy by performing one-shot pruning on spatial-temporal message-passing graphs, yielding token-economic topologies while defending against adversarial attacks. (2) \textbf{Agent-Dropout}~\cite{wang2025agentdropout}: Inspired by dynamic role adjustment in management theory, identifies redundant agents and communications across different rounds by optimizing adjacency matrices, eliminating them to enhance token efficiency and task performance. (3) \textbf{G-Designer}~\cite{zhang2025g}: Employs a variational graph auto-encoder to encode agents and task-specific virtual nodes, dynamically designing task-adaptive communication topologies customized for each task's difficulty and requirements. (4) \textbf{EIB-LEARNER}~\cite{shen2025understanding}: Uses a causal framework to analyze error propagation patterns and designs topologies by fusing connectivity patterns from both dense and sparse graphs, balancing error suppression with beneficial information diffusion. (5) \textbf{ARG-Designer}~\cite{li2025assemble}: Autoregressively constructs communication topologies from scratch by sequentially adding individual agents with task-specific role assignments and establishing pairwise connections, bypassing the need for predefined template graphs.

\section{Parameter Sensitivity Analysis}
\label{sec:parameter_sensitivity_analysis}
The Conditional Information Bottleneck (CIB) is governed by two key hyperparameters: $\beta_g$ and $\beta_e$, which control the strength of the KL divergence penalty during group prediction and edge prediction, respectively. We investigate the sensitivity of \method to these parameters on the MMLU dataset.

\begin{figure}[!ht]
\centering
\begin{subfigure}[t]{0.48\columnwidth}
\centering
\includegraphics[width=\linewidth]{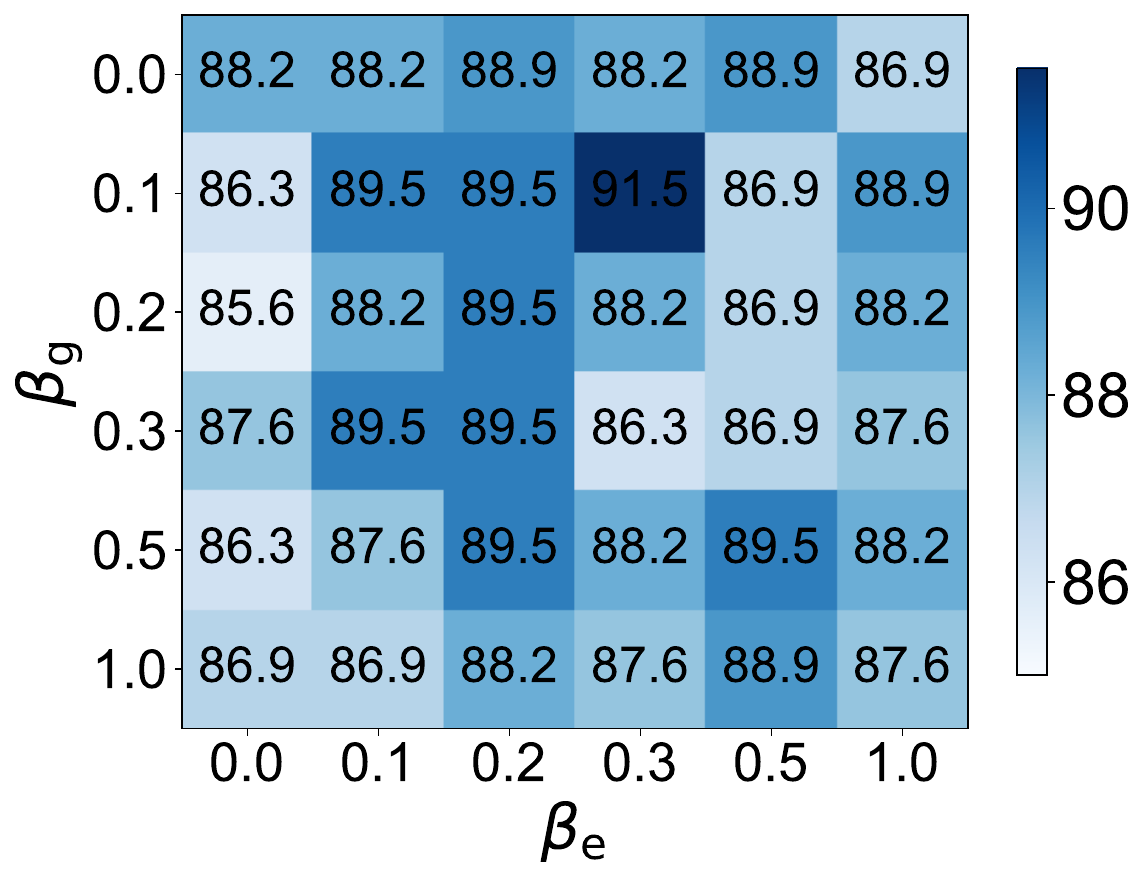}
\caption{Accuracy (\%).}
\end{subfigure}
\hspace{-2mm}
\begin{subfigure}[t]{0.48\columnwidth}
\centering
\includegraphics[width=\linewidth]{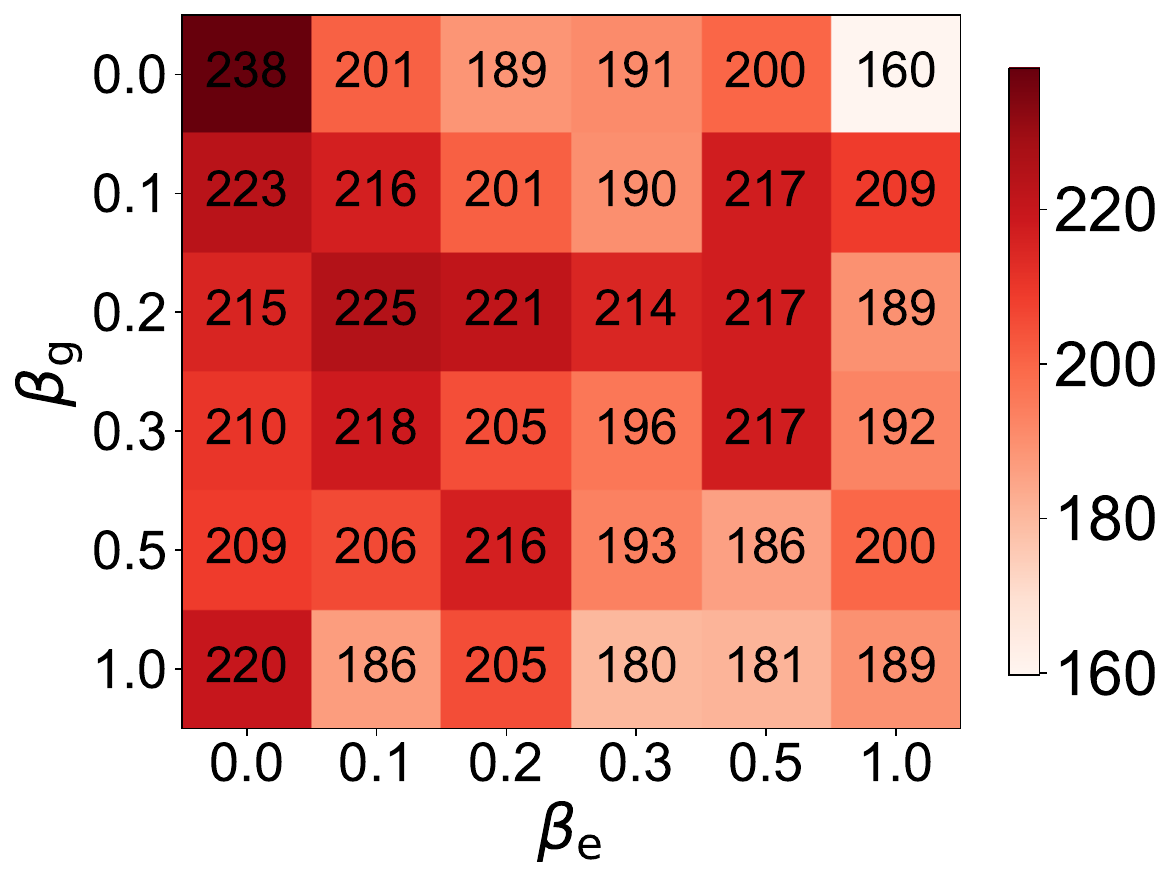}
\caption{Token cost (K).}
\end{subfigure}
\caption{Heatmap of model performance with varying $\beta_g$ and $\beta_e$ on the MMLU dataset.}
\label{fig:heatmap}
\end{figure}

As illustrated in Figure~\ref{fig:heatmap}, the model's performance is highly sensitive to the bottleneck strength. Setting $\beta$ too low (e.g., $\beta_e = 0.0$) renders the bottleneck ineffective, allowing historical noise to pass through and resulting in over-connected graphs with high token costs and lower accuracy. Conversely, setting $\beta$ too high (e.g., $\beta_e > 0.5$) forces the model to aggressively compress the features, discarding necessary historical context and leading to under-connected, poorly coordinated topologies.

Interestingly, the optimal configuration occurs in a highly asymmetric region ($\beta_g \approx 0.0$ and $\beta_e \approx 0.3$). This indicates that edge prediction benefits significantly more from compression than group prediction. This asymmetry aligns with our intuition: determining \emph{which} group to generate next (group prediction) requires full access to the rich historical context to avoid duplicating roles. However, determining \emph{how} these groups connect (edge prediction) is highly susceptible to the redundant connectivity patterns common in prior generative models. The CIB layer effectively filters out this structural noise, ensuring that edges are formed only when task-relevant information must be routed between groups.

\section{Implementation Details}
\label{sec:implementation_details}
\paragraph{Training.} During the heuristic exploration described in Section~\ref{sec:training}, candidate topologies are drawn from three structural families: Chain, Star, and FullConnected. We train with AdamW (learning rate $1 \times 10^{-4}$, weight decay $1 \times 10^{-3}$, gradient clipping $1.0$) for $E = 100$ epochs with a batch size of 40. The CIB bottleneck strength undergoes linear warm-up over the first $E_{\text{warm}} = 10$ epochs. Notably, only $B \in \{40, 60\}$ training queries per dataset suffice, because the generator learns \emph{structural collaboration patterns} (which group types co-occur and how they connect) rather than task-specific reasoning, which is handled entirely by the underlying LLMs.

\paragraph{Architecture.} All modules use hidden dimension $h = 256$. The task encoder FFN projects $d \rightarrow h \rightarrow d$ with ReLU. The historical aggregation GRU, group prediction GRU, and edge prediction GRU are each single-layer with input and hidden size $d$. For edge prediction, the concatenated feature $\mathbf{x}_{\text{edge}}^{(i,t)} \in \mathbb{R}^{3d}$ is projected via a two-layer MLP ($3d \rightarrow h \rightarrow d$, ReLU), processed by the edge GRU, then passed through the CIB layer. The final edge classification head maps $d \rightarrow h \rightarrow 1$ with sigmoid output. During inference, edges are sampled via Bernoulli; if no edge is sampled for a newly added group, the highest-probability edge is added to guarantee connectivity.

\subsection{LLM Prompt Templates}
We use GPT-4o with temperature 0.7 to generate diverse (query, topology) pairs. The system prompt instructs the LLM to generate queries spanning multiple domains (code implementation, mathematical reasoning, multiple-choice questions, knowledge-based Q\&A, medical consultation, psychology) and propose suitable MAS configurations using roles from our group pool and topologies from \{Chain, Star, FullConnected\}. A representative example output is shown below:

\noindent\textbf{Example Output:}
{\small
\begin{verbatim}
{"query": "def odd_count(lst)...", "agent_count": 4,
 "roles": ["solver group", "verifier group",
           "knowledge group", "coder group"],
 "topology": "Chain", "edges": "0->1 1->2 2->3"}
\end{verbatim}
}

\subsection{Group Role Prompts}
\label{sec:group_role_prompts}
Our group role pool contains 16 specialized groups for GSM8K/MultiArith/SVAMP/AQuA. Representative group generation and prompts include:

\begin{tcolorbox}[colback=gray!10, colframe=gray!50, title=Role-to-Group Consolidation Instructions (Abbreviated)]
  \small\raggedright
  \textbf{Base Role Pool:} 4 specialized roles --- Math Solver (step-by-step computation), Mathematical Analyst (symbolic reasoning \& decomposition), Programming Expert (Python implementation), Inspector (validation \& verification).
  
  \textbf{Group Prompt Design:} Each group's system prompt describes all internal roles, their sequential stages, connection semantics, and collaboration rationale, enabling a single LLM call to simulate the entire multi-role interaction.
  
  \textbf{Consolidation Topology Categories:}
  
  {\footnotesize
  Single: Independent standalone group.
  
  Two-Stage Chain: Pairwise pipelines, e.g., Analyst$\rightarrow$Solver.
  
  Three-Stage Chain: Three-role pipelines, e.g., Analyst$\rightarrow$Programming$\rightarrow$Inspector.
  
  Star: Many-to-one cross-validation, e.g., (Programming, Solver)$\rightarrow$Inspector.
  
  FullConnected: Fully connected DAG, e.g., Analyst$\rightarrow$Solver, Analyst$\rightarrow$Inspector, Solver$\rightarrow$Inspector.
  }

  \textbf{DAG Constraint:} All group-internal topologies are Directed Acyclic Graphs. Information flows strictly from upstream roles to downstream roles.
  
  \textbf{Coverage:} 16 distinct group configurations derived from 4 base roles $\times$ 5 topology patterns, ensuring comprehensive coverage of collaboration paradigms from independent reasoning to multi-stage validated pipelines.
  \end{tcolorbox}

\begin{tcolorbox}[colback=blue!5, colframe=blue!50, title=Math Solver Group (Chain)]
\small
You operate as a collaborative two-stage chain connecting Mathematical Analyst and Math Solver roles.

Role 1 (Mathematical Analyst): You analyze the problem structure using symbolic variables, identifying key relationships, formulas, and computational steps in abstract form.

Role 2 (Math Solver): You receive the symbolic analysis and substitute actual values from the problem to compute the final numerical answer step by step.
The collaboration transforms abstract mathematical structure into concrete numerical solutions through sequential processing.
The connection structure: Analyst $\rightarrow$ Solver enables structured problem decomposition followed by precise calculation.
\end{tcolorbox}

\begin{tcolorbox}[colback=blue!5, colframe=blue!50, title=Code Solver Group (Chain)]
\small
You operate as a collaborative three-stage chain connecting Mathematical Analyst, Programming Expert, and Inspector roles.

Role 1 (Mathematical Analyst): You analyze the problem structure using symbolic variables, identifying key relationships, formulas, and the computational approach.

Role 2 (Programming Expert): You receive the symbolic analysis and implement it as Python code with well-structured functions that algorithmically solve the problem.

Role 3 (Inspector): You receive the code and verify its correctness against the original analysis, checking logic flow, calculation accuracy, and code-analysis correspondence.

The collaboration transforms abstract analysis into verified executable code through sequential processing with quality assurance.
The connection structure: Analyst $\rightarrow$ Programming Expert $\rightarrow$ Inspector enables structured analysis, followed by implementation, followed by validation.
\end{tcolorbox}

\begin{tcolorbox}[colback=blue!5, colframe=blue!50, title=Solution Verifier Group (Star)]
\small
You operate as a collaborative star structure where Mathematical Analyst and Math Solver both feed into Inspector for comprehensive validation.

Role 1 (Mathematical Analyst): You provide the symbolic analysis and mathematical framework of the problem.

Role 2 (Math Solver): You provide the step-by-step numerical solution with intermediate calculations.

Role 3 (Inspector): You receive both the analytical framework and the numerical solution, verifying that the solution correctly follows the established analysis.

The collaboration enables validation of the solution against its analytical foundation, ensuring logical consistency.
The connection structure: Analyst $\rightarrow$ Inspector AND Solver $\rightarrow$ Inspector enables framework-based solution verification.
\end{tcolorbox}

\begin{tcolorbox}[colback=blue!5, colframe=blue!50, title=Full Validation Group (Full Connected)]
\small
You operate as a collaborative Full Connected structure where all three roles communicate with downstream roles in a directed acyclic graph.

Role 1 (Mathematical Analyst): You analyze the problem structure, providing symbolic framework to both Math Solver (for computation) and Inspector (for validation criteria).

Role 2 (Math Solver): You receive the analysis from Analyst, compute the solution, and pass your results to Inspector for verification.

Role 3 (Inspector): You receive the analytical framework from Analyst AND the solution from Solver, performing comprehensive validation against both.

The collaboration enables maximum information flow, where the Inspector validates the solution against both the original analysis and the computation process.
The connection structure: Analyst $\rightarrow$ Solver, Analyst $\rightarrow$ Inspector, Solver $\rightarrow$ Inspector forms a complete validation triangle.
\end{tcolorbox}

\end{document}